\documentclass[times,twocolumn,final,authoryear]{elsarticle}

\usepackage{ycviu}
\usepackage{framed,multirow}

\usepackage{amssymb}
\usepackage{latexsym}
\usepackage{amsmath}
\usepackage{breqn}
\usepackage{subfig}
\usepackage{stfloats}

\usepackage{url}
\usepackage{xcolor}
\definecolor{newcolor}{rgb}{.8,.349,.1}

\journal{}

\begin{document}

\begin{frontmatter}

\title{Depth-aware Blending of Smoothed Images for Bokeh Effect Generation}

\author[1]{Saikat \snm{Dutta}\corref{cor1}} 
\cortext[cor1]{Corresponding author}
\ead{saikat@smail.iitm.ac.in}

\address[1]{Indian Institute of Technology Madras, Chennai, PIN-600036, India}

\received{1 May 2013}
\finalform{10 May 2013}
\accepted{13 May 2013}
\availableonline{15 May 2013}
\communicated{S. Sarkar}

\begin{abstract}
Bokeh effect is used in photography to capture images where the closer objects look sharp and everything else stays out-of-focus. Bokeh photos are generally captured using Single Lens Reflex cameras using shallow depth-of-field. Most of the modern smartphones can take bokeh images by leveraging dual rear cameras or a good auto-focus hardware. However, for smartphones with single-rear camera without a good auto-focus hardware, we have to rely on software to generate bokeh images. This kind of system is also useful to generate bokeh effect in already captured images. In this paper, an end-to-end deep learning framework is proposed to generate high-quality bokeh effect from images. The original image and different versions of smoothed images are blended to generate Bokeh effect with the help of a monocular depth estimation network. The proposed approach is compared against a saliency detection based baseline and a number of approaches proposed in AIM 2019 Challenge on Bokeh Effect Synthesis. Extensive experiments are shown in order to understand different parts of the proposed algorithm. The network is lightweight and can process an HD image in 0.03 seconds. This approach ranked second in AIM 2019 Bokeh effect challenge-Perceptual Track.
\end{abstract}


\end{frontmatter}


\section{Introduction}
Depth-of-field effect or Bokeh effect is often used in photography to generate aesthetic pictures. Bokeh images basically focus on a certain subject and out-of-focus regions are blurred. Bokeh images can be captured in Single Lens Reflex cameras using high aperture. In contrast, most smartphone cameras have small fixed-sized apertures that can not capture bokeh images. Many smartphone cameras with dual rear cameras can synthesize bokeh effect. Two images are captured from the cameras and stereo matching algorithms are used to compute depth maps and using this depth map, depth-of-field effect is generated. Smartphones with good auto-focus hardware e.g. iPhone7+ can generate depth maps which helps in rendering Bokeh images. However, smartphones with single camera that don't have a good auto-focus sensor have to rely on software to synthesize bokeh effect.

Also, already captured images can be post-processed to have Bokeh effect by using this kind of software. That is why generation of synthetic depth-of-field or Bokeh effect is an important problem in Computer Vision and has gained attention recently. Most of the existing approaches\citep{shen2016automatic,wadhwa2018synthetic,xu2018rendering} work on human portraits by leveraging image segmentation and depth estimation. However, not many approaches have been proposed for bokeh effect generation for images in the wild. Recently, \cite{purohit2019depth} proposed an end-to-end network to generate bokeh effect on random images by leveraging monocular depth estimation and saliency detection.

In this paper, one such algorithm is proposed that can generate Bokeh effect from diverse images. The proposed approach relies on a depth-estimation network to generate weight maps that blend the input image and different smoothed versions of the input image. The generated bokeh images by this algorithm are visually pleasing. The proposed approach ranked 2nd in AIM 2019 challenge on Bokeh effect Synthesis- Perceptual Track\citep{ignatov2019aim}.

\section{Related Work}
\subsection{Monocular Depth Estimation}
Depth estimation from a single RGB image is a significant problem in Computer Vision with a long range of applications including robotics, augmented reality and autonomous driving. Recent advances in deep learning have helped in the progress of monocular depth estimation algorithms. Supervised algorithms rely on ground truth Depth data captured from depth sensors. \cite{svs} formulated monocular depth estimation as a combination of two sub-problems: view synthesis and stereo matching. View synthesis network creates another view of the given RGB image which is utilized by the stereo matching network to estimate the depth map. \cite{chen2016single} train a multi-scale deep network that predicts pixelwise metric depth with relative-depth annotations from images in the wild. \cite{depth_style_transfer} trains a depth estimation network on large amount of synthetic data and use image style transfer to map real world images to synthetic images. \cite{monodepth} poses Monocular Depth Estimation as an image reconstruction problem. They train a Convolutional Neural Network on easy-to-obtain binocular stereo data rather than training on depth ground truth and introduce a novel loss function that checks left-right consistency in generated disparity maps. \cite{monodepth2} introduces a minimum reprojection loss, an auto masking loss and a full-resolution multi-scale sampling method to train a deep network in self-supervised way. Their minimum reprojection loss handles occlusion in a monocular video and auto-masking loss helps network to ignore confusing and stationary pixels. \cite{li2018megadepth} created a training dataset using Structure from Motion and Multi-view Stereo methods. They trained three different networks using scale-invariant reconstruction and gradient matching loss and ordinal depth loss which achieves good generalization performance.

\begin{figure*}[!h!t]
\includegraphics[width=\textwidth]{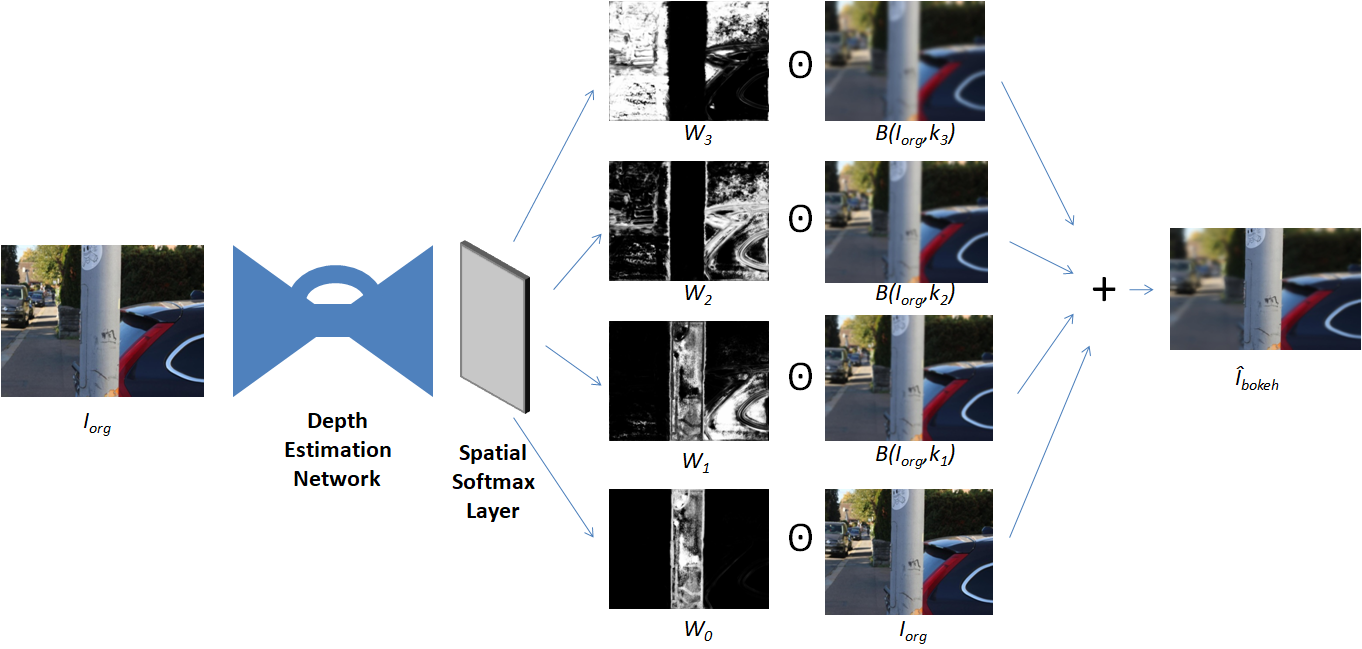}
\caption{Pipeline of the proposed method.}
\label{proposed_method}
\end{figure*}

\begin{figure*}[!ht]
    \centering
    \includegraphics[width=0.8\textwidth]{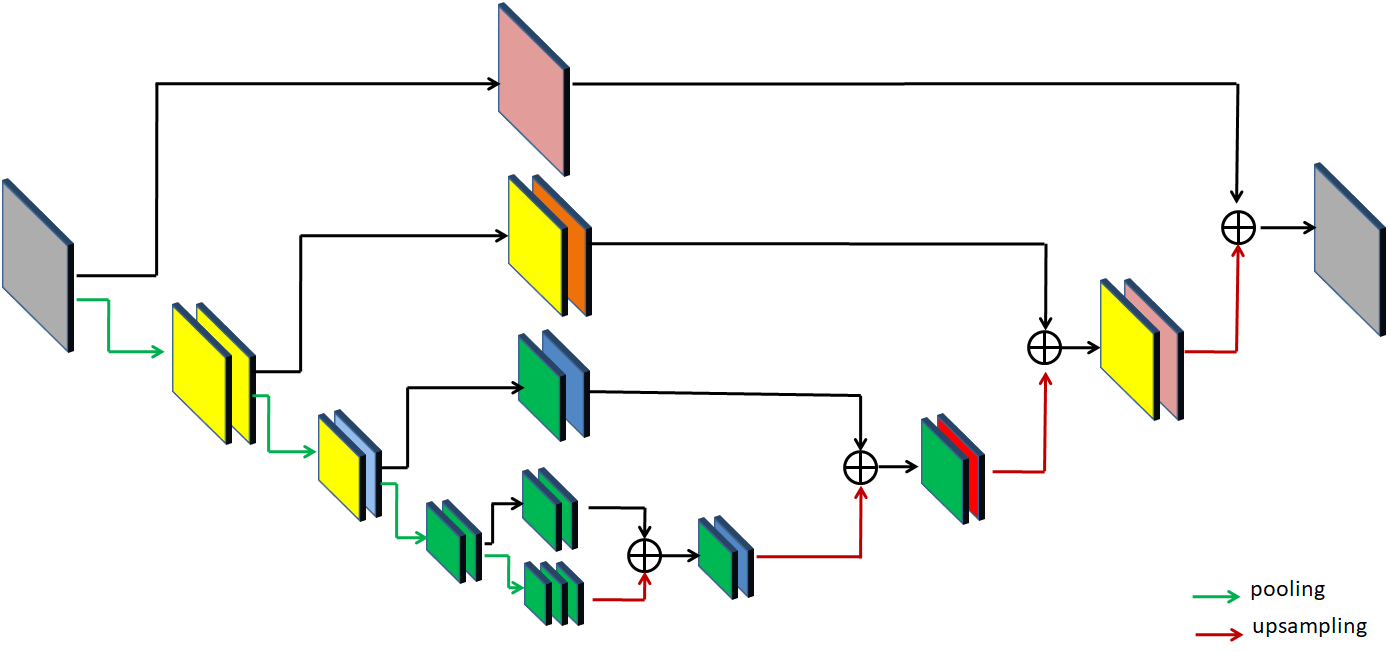}
    \caption{MegaDepth Architecture. Different colors denote different convolutional blocks. The last convolutional block is modified in the proposed system.}
    \label{megadepth_arch}
\end{figure*}

\subsection{Depth Effect in Images}
Early work in Depth-of-field generation in images uses automatic portrait segmentation. \cite{shen2016automatic} used a FCN\citep{fcn} network to segment the foreground in the images and use the foreground for different kind of background editing such as background replacement, BW one color and depth-of-field effect. \cite{wadhwa2018synthetic} uses a neural network to segment person and its accessories in an image to generate a foreground mask in the case of person images. Then they generate dense depth maps using a sensor with dual-pixel auto-focus hardware and use this depth map along with foreground mask (if available) for depth-dependent rendering of shallow depth-of-field images. \cite{xu2018rendering} too uses both learning based and traditional techniques to synthesize depth-of-field effect. They use off-the-shelf networks for portrait segmentation and single image depth estimation and use SPN\citep{liu2017learning} to refine the estimated segmentation and depth maps. They split the background into different layers of depth using CRF\citep{zheng2015conditional} and render the bokeh image. For fast rendering, they also learn a spatially varying RNN filter\citep{liu2016learning}) with CRF-rendered bokeh result as ground truth.

\cite{purohit2019depth} proposed an end-to-end network for Bokeh Image Synthesis. First, they compute depth map and saliency map using pretrained monocular depth estimation and saliency detection networks respectively. These maps and the input image are fed to a densely-connected hourglass architecture with pyramid pooling module. This network generates spatially-aware dynamic filters to synthesize corresponding bokeh image.

In this paper, an end-to-end deep learning approach is proposed for Bokeh effect rendering. In this approach, the corresponding bokeh image is thought to be a weighted sum of different smoothed versions of the input image. The spatially varying weights are learned with the help of a pretrained monocular depth estimation network by modifying the last convolutional layer. Compared to some other methods in the literature, this approach doesn't rely on foreground or salient region segmentation, which makes the proposed algorithm lightweight. The proposed method generates good bokeh effect rendering both qualitatively and quantitatively. This approach came 2nd in Perceptual track of AIM 2019 challenge on Bokeh effect synthesis.

\begin{figure*}[!h!t]
    \centering
    \includegraphics[width=\textwidth]{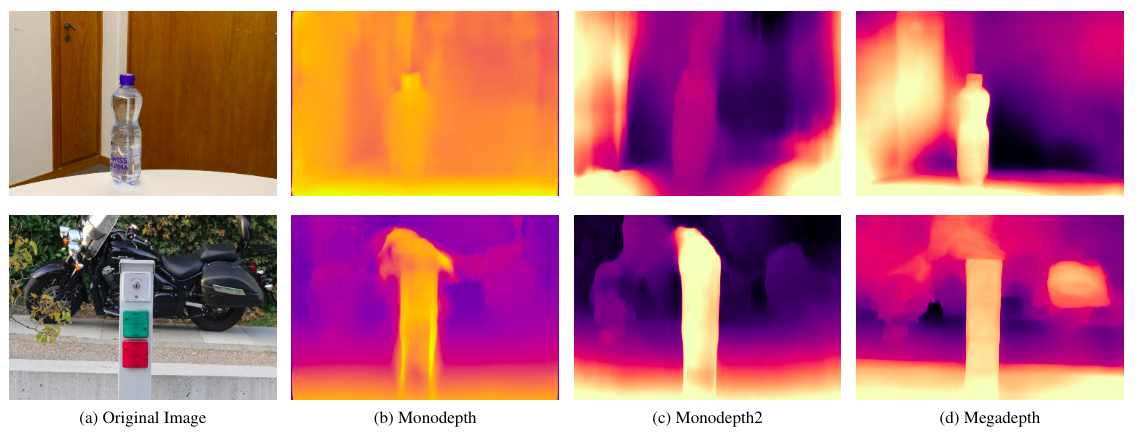}
    \caption{Monocular depth estimation result comparison among different state-of-the-art models.}
\label{depth_comparison}
\end{figure*}

\section{Proposed Methodology}

\subsection{Depth Estimation Network}
Megadepth\citep{li2018megadepth} is used as Monocular Depth estimation network in this work. The authors use an hourglass architecture which was originally proposed in \cite{chen2016single}. The architecture is shown in Figure-\ref{megadepth_arch}. The encoder part of this network consists of a series of convolutional modules (which is a variant of inception module) and downsampling. In the decoder part, there is a series of convolutional modules and upsampling with skip connections that add back features from higher resolution in between.

Megadepth was trained using a large dataset collected from internet using structure-from-motion and multi-view stereo methods. Megadepth works quite well in a wide variety of datasets such as KITTI\citep{geiger2012we}, Make3D\citep{saxena2006learning} and DIW\citep{chen2016single}. This generalization ability of Megadepth makes it suitable for bokeh effect rendering for diverse scenes. Figure-\ref{depth_comparison} shows predicted depth maps by Megadepth and other state-of-the-art models. We can see that Megadepth produces better quality depth estimation compared to other models.

\subsection{Bokeh effect rendering}
The depth-of-field image can be formulated as a weighted sum of the original image and differently smoothed versions of original image. These different smoothed images can be generated using different size of blur kernels. Concretely, intensity value of each pixel position in the bokeh image can be thought of as a weighted sum of intensity values from original image and smoothed images in that pixel position. Hence, the generated bokeh image is given by,
\begin{equation}
    \hat{I}_{bokeh} = W_0\odot I_{org} + \sum_{i=1}^{n} W_i \odot B(I_{org},k_i)
\label{bokeh_gen_equ}    
\end{equation}
where $I_{org}$ is the original image, $B(I_{org},k_i)$ is image $I_{org}$ smoothed by blur kernel of size $k_i\times k_i$ and $\odot$ stands for elementwise multiplication, such that for each pixel position $(x,y)$,  
\begin{equation}
    \sum_{i=0}^{n} W_{i}[x,y] = 1
\end{equation}
In the proposed approach, the weights 
$W_i$ are predicted with the help of a neural network. This can be achieved by modifying the last layer of the pretrained depth estimation network. Specifically, in the last layer a convolutional layer with $3\times3$ kernel is used along with spatial softmax activation to learn the weights. The proposed approach is summarized in Figure-\ref{proposed_method}.

In the experiments, $n$ is chosen to be 3 i.e. three different smoothed images are used. The smoothed images were obtained by applying Gaussian Blur with Blur kernels of size $25\times25$, $45\times 45$ and $75\times 75$. The effect of using less number of kernels and kernels of different size are discussed in the following section.

\section{Experiments}
\subsection{System Configuration}
The codes were written in Python and Pytorch(\citep{pytorch}) is used as the deep learning framework. The models were trained on a machine with Intel Xeon 2.40 GHz processor, 64 GB RAM and NVIDIA GeForce TITAN X GPU card with approximately 12GB of GPU memory.

\subsection{Dataset Description}
\begin{figure*}[!h!t]
    \centering
    \includegraphics[width=\textwidth]{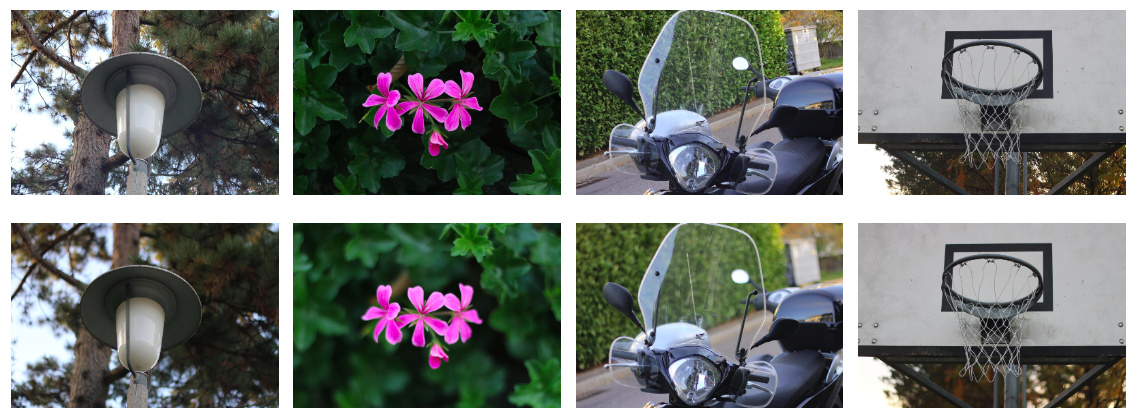}
    \caption{Diverse images from ETH Zurich Bokeh Dataset. Top row contains bokeh-free images and bottom row contains corresponding bokeh images.}
\label{snapshot}
\end{figure*}

We use ETH Zurich Bokeh dataset\citep{ignatov2019aim}, which was used in AIM 2029 Bokeh Effect Synthesis Challenge. This dataset contains 4893 pairs of bokeh and bokeh-free images. Training set contains 4493 pairs whereas Validation and Testing datasets contain 200 pairs each. Validation and Testing Bokeh images are not publicly available at this moment. This dataset contains a variety of outdoor scenes. The normal images were captured using narrow aperture ($f/16$) and a high aperture ($f/1.8$) was used to generate the Bokeh effect in images. The images in this dataset have resolution of around $1024 \times 1536$ pixels. A snapshot of this dataset is given in Figure-\ref{snapshot}. 

The training set of ETH Zurich Bokeh Dataset is divided to a train set containing 4400 pairs and a validation set containing 294 pairs for experiments in this paper. This validation set is denoted as \textit{Val294}.

\subsection{Training}
\subsubsection{Loss functions}
\textbf{Reconstruction loss:} $l_1$ loss is used as reconstruction loss to guide the network to generate images with pixel values close to ground truth. Reconstruction loss is given by,
\begin{equation}
    l_r = \left\|\hat{I}_{bokeh}-I_{bokeh}\right\|_{1}
\end{equation}
\textbf{Perceptual loss:} Negative SSIM\citep{ssim_paper} loss is used to improve perceptual quality of the generated images. Perceptual loss is given by,
\begin{equation}
    l_p = -SSIM(\hat{I}_{bokeh}, I_{bokeh})
\end{equation}
\subsubsection{Training Strategy}
The model is trained in 3 phases. In phase-1, the model is trained on smaller resolution images to save training time. Each image from training set 
is resized to a resolution of $384 \times 512 $ and the network is trained using reconstruction loss. In phase-2, the network is trained using images of resolution $768 \times 1024 $ with reconstruction loss. In phase-3, the model is further fine-tuned with perceptual loss $l_p$.

\subsubsection{Other training details} Adam optimizer(\citep{kingma2014adam}) is used to train the network. The values of $\beta_1$ and $\beta_2$ are chosen to be $0.9$ and $0.999$ respectively. The initial learning rate is set to be 1e-3 and gradually decreased to 1e-5. Each image in the training set is horizontally and vertically flipped to augment the dataset. 

\section{Results}
\subsection{Testing Strategy}
During testing, the input image is first resized to the original dimension in which the network was trained($384 \times 512 $ for Phase-1 and $768 \times 1024 $ for Phase-2 and Phase-3) and then passed to the network. The synthesized image is then scaled back to the input image resolution using bilinear interpolation.

\subsection{Evaluation metrics}
Both fidelity and perceptual metrics are used to evaluate the model's performance. Fidelity measures include Peak Signal-to-Noise Ratio(PSNR) and Structural Similarity Index(SSIM)\citep{ssim_paper}. LPIPS\citep{LPIPS} is used to measure perceptual quality of the generated results.

\begin{figure*}[!th]
    \centering
    \includegraphics[width=\textwidth]{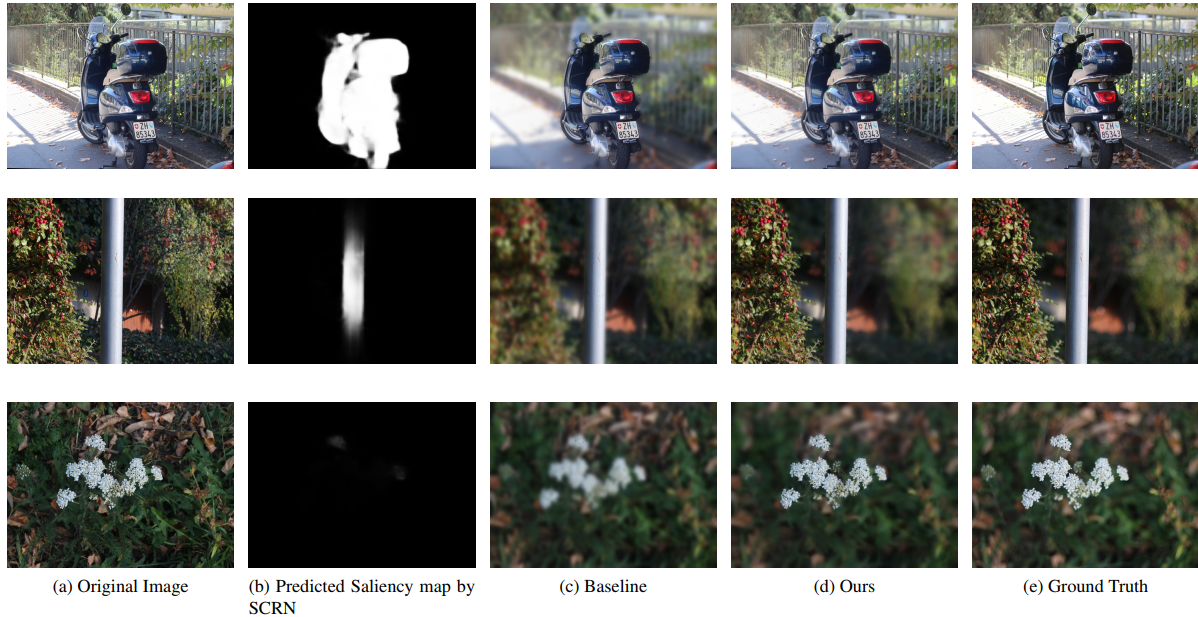}
    \caption{Baseline results from Saliency detection and comparison against the proposed model.}
    \label{scrn}
\end{figure*}

\subsection{Quantitative and Qualitative Results}
\textbf{Baseline.} Bokeh effect can be generated by simply segmenting the foreground in the image and blurring the rest of the image. One such system is used as baseline in this paper. A state-of-the-art saliency detection method, Stacked Cross Refinement Network(SCRN)\citep{wu2019stacked} is used for segmenting the foreground in the input image. The background is blurred using $75 \times 75 $ Gaussian kernel. As SCRN generates soft saliency maps, bokeh images are generated using Equ-\ref{bokeh_gen_equ}.

Quantitative comparison between Saliency detection based baseline and the proposed method is shown in Table-\ref{scrn}. We can see that the proposed approach performs significantly better than the baseline on all three metrics. Figure-\ref{scrn} shows saliency maps generated by SCRN and corresponding bokeh effect rendering by baseline and the proposed model. We can see that the performance of the baseline is limited to accuracy of saliency detection which leads to artifacts near edges and incorrect blurring of foreground in rendered images.

\begin{table}[!h]
\begin{tabular}{|c|c|c|c|}
\hline
               & PSNR $\uparrow$ & SSIM $\uparrow$ & LPIPS $\downarrow$ \\ \hline
Baseline(SCRN) &     22.53                         &       0.8330                       &        0.3251                         \\ \hline
Proposed       &       \textbf{23.45}                       &         \textbf{0.8675}                     &          \textbf{0.2463}                       \\ \hline
\end{tabular}
\centering
\caption{Quantitative comparison between Baseline and the proposed approach on Val294 set.}
\label{tab_scrn}
\end{table}

\subsection{Visualization}
\begin{figure*}[!h!t!]
    \centering
    \includegraphics[width=\textwidth]{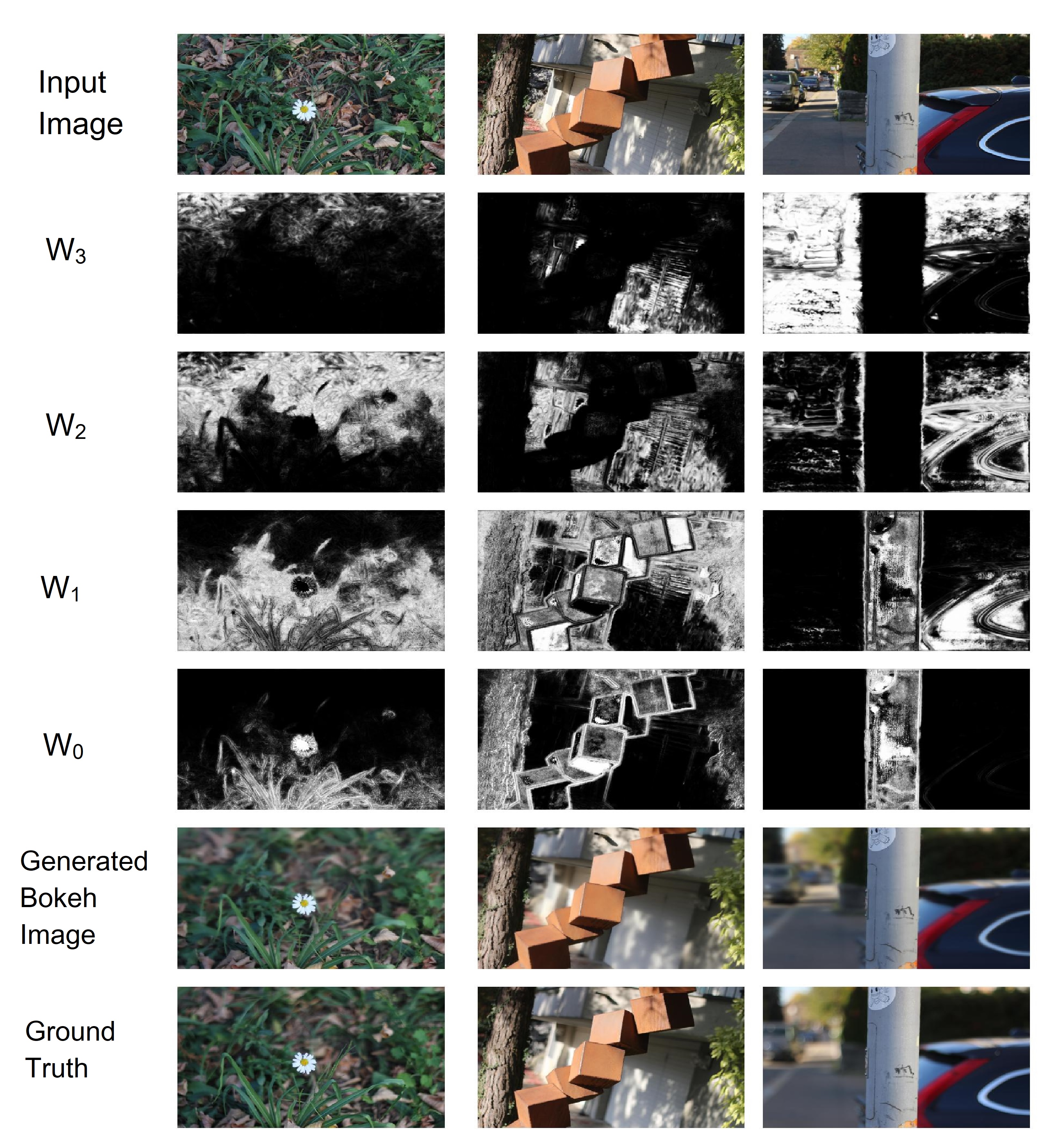}
    \caption{Visualization of weight maps generated by proposed model. Closer objects have larger weights in $W_0$ and $W_1$ compared to $W_2$ and $W_3$.}
    \label{wt_vis}
\end{figure*}

Examples of generated bokeh images and the corresponding predicted weight maps by our network are shown in Figure-\ref{wt_vis}. We can see that, image regions that are closer have higher weights in $W_0$ and $W_1$ compared to $W_2$ and $W_3$. This allows the model to generate blur effect that is varying with respect to depth i.e. closer objects look sharp and distant regions get more blurred with increasing depth in the rendered bokeh image.

\subsection{Ablation Study}
\begin{figure*}[!h!t!p]
     \centering
     \includegraphics[width=\textwidth]{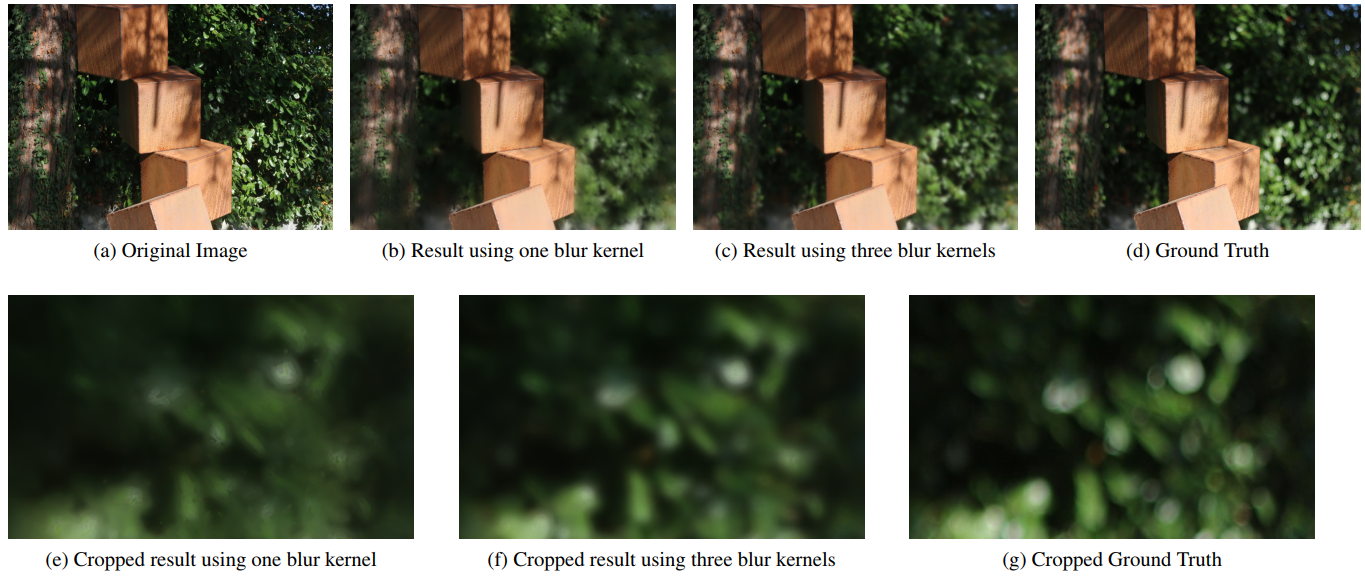}
     \caption{Qualitative Comparison between using different number of blur kernels.}
     \label{less_kernels}
 \end{figure*}

\textbf{Less number of kernels vs. more number of kernels.}
The main intuition behind using
a number of kernels is to try to synthesize bokeh images in a way that regions that are closer to the camera have no or less blur effect whereas distant objects have more blur effect. Using less number of kernels will guide network to learn lesser number of such blur levels in the image. Using more number of blur kernels helps the network two generate smoother background in synthesized bokeh images. Table-\ref{tab_less_kernels} shows quantitative comparison between using just one blur kernel of size $75$ and the proposed approach. We can observe in Figure-\ref{less_kernels} using more number of kernels produces more visually pleasing background.


\begin{table}[!h]
\begin{tabular}{|c|c|c|c|}
\hline
                                                                                         & PSNR $\uparrow$ & SSIM $\uparrow$ & LPIPS $\downarrow$ \\ \hline
\begin{tabular}[c]{@{}c@{}}One kernel\\ ($k_1 = 75$)\end{tabular}                        &    23.17           &      0.8613         &     0.2506             \\ \hline
\begin{tabular}[c]{@{}c@{}}Three kernels\\ ($k_1 = 25, k_2 = 45, k_3 = 75$)\end{tabular} &    \textbf{23.45}           &       \textbf{0.8675}        &   \textbf{0.2463}               \\ \hline 
\end{tabular}
\centering
\caption{Quantitative comparison between using different number of kernels on Val294 set.}
\label{tab_less_kernels}
\end{table}

\begin{figure*}[!h!t]
\centering
\subfloat[Original Image]{\includegraphics[width = 0.24\linewidth,height=1.2in]{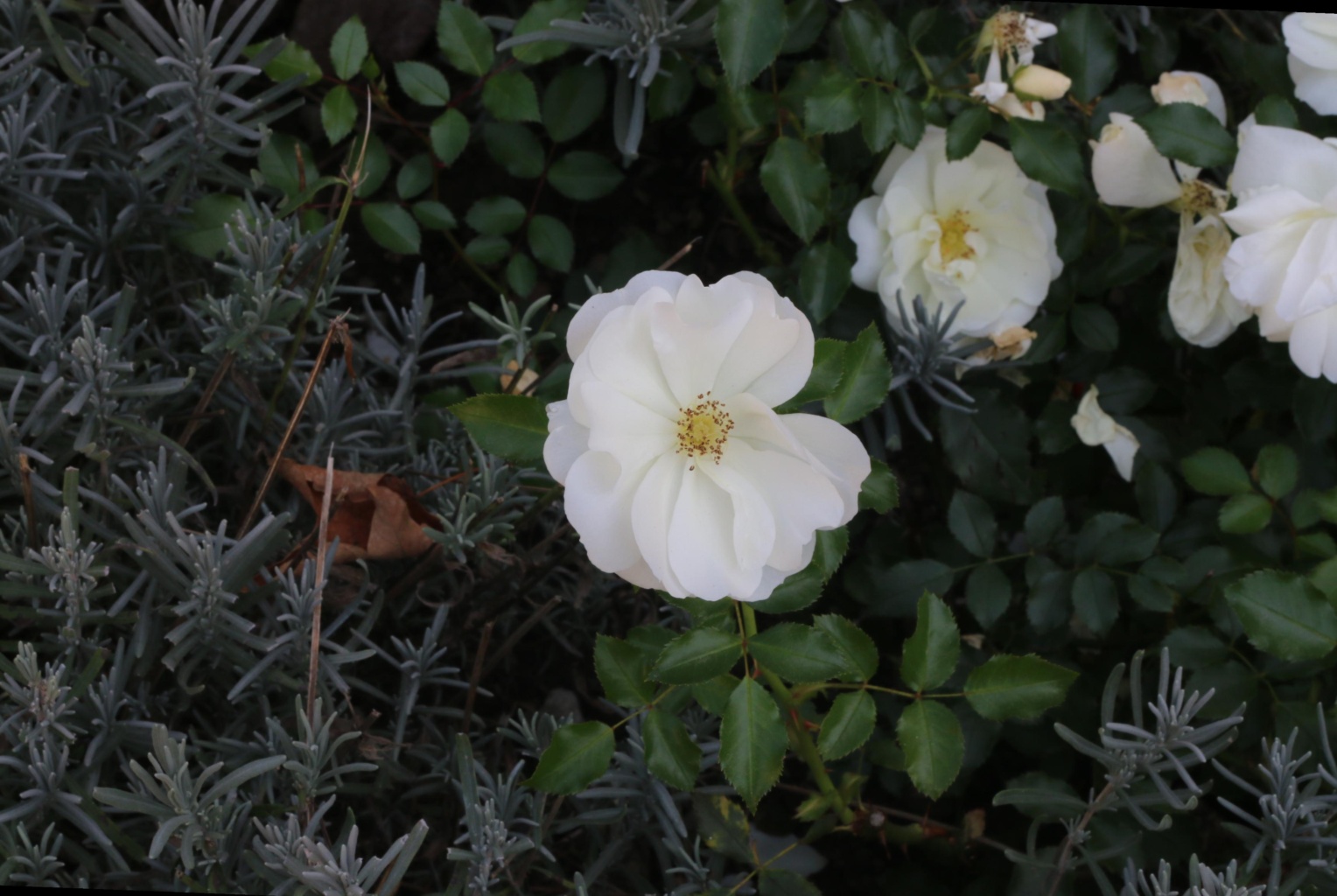}} \hfill
 \subfloat[Bokeh using small kernels]{\includegraphics[width = 0.24\linewidth,height=1.2in]{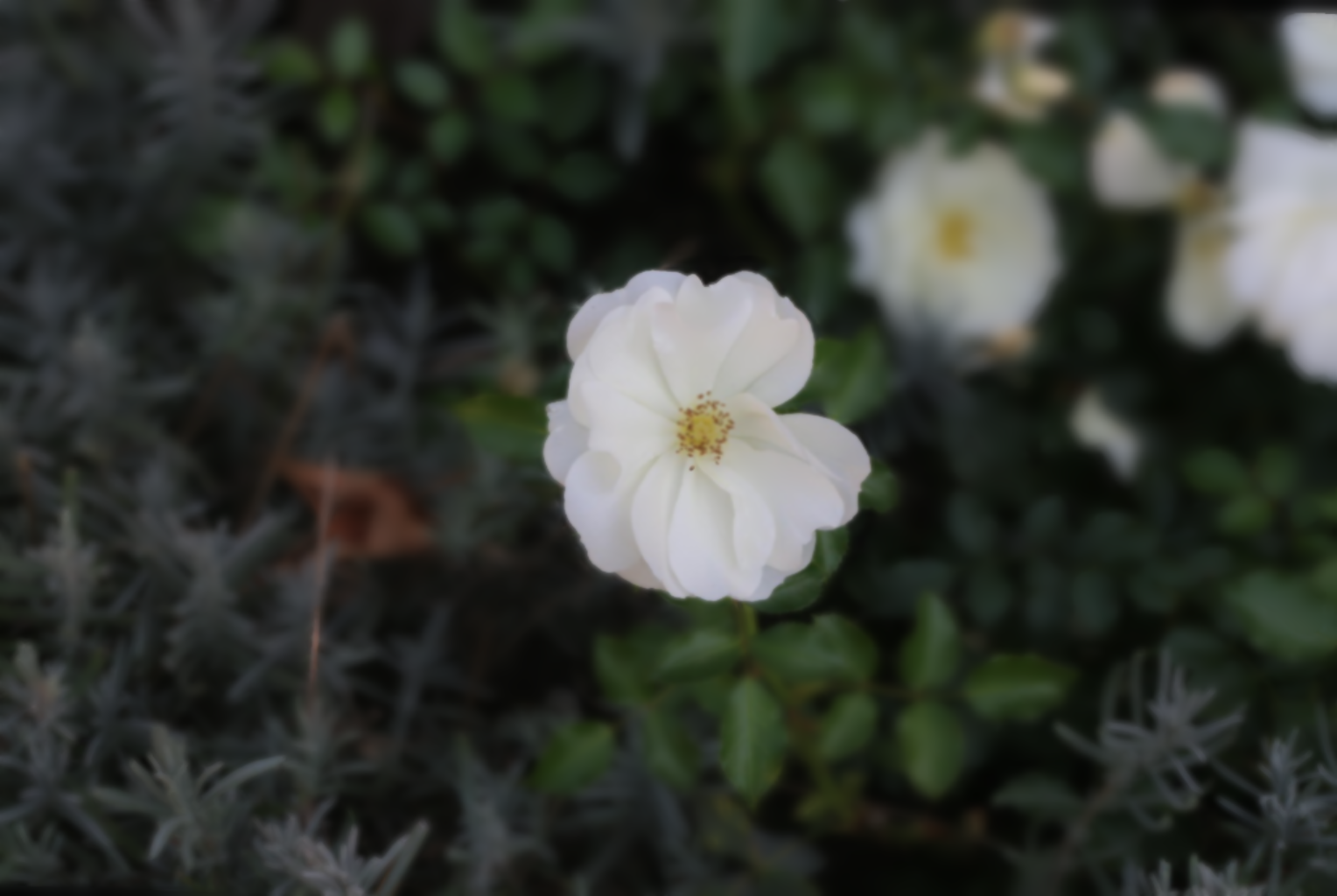}} \hfill
 \subfloat[Bokeh using bigger kernels]{\includegraphics[width = 0.24\linewidth,height=1.2in]{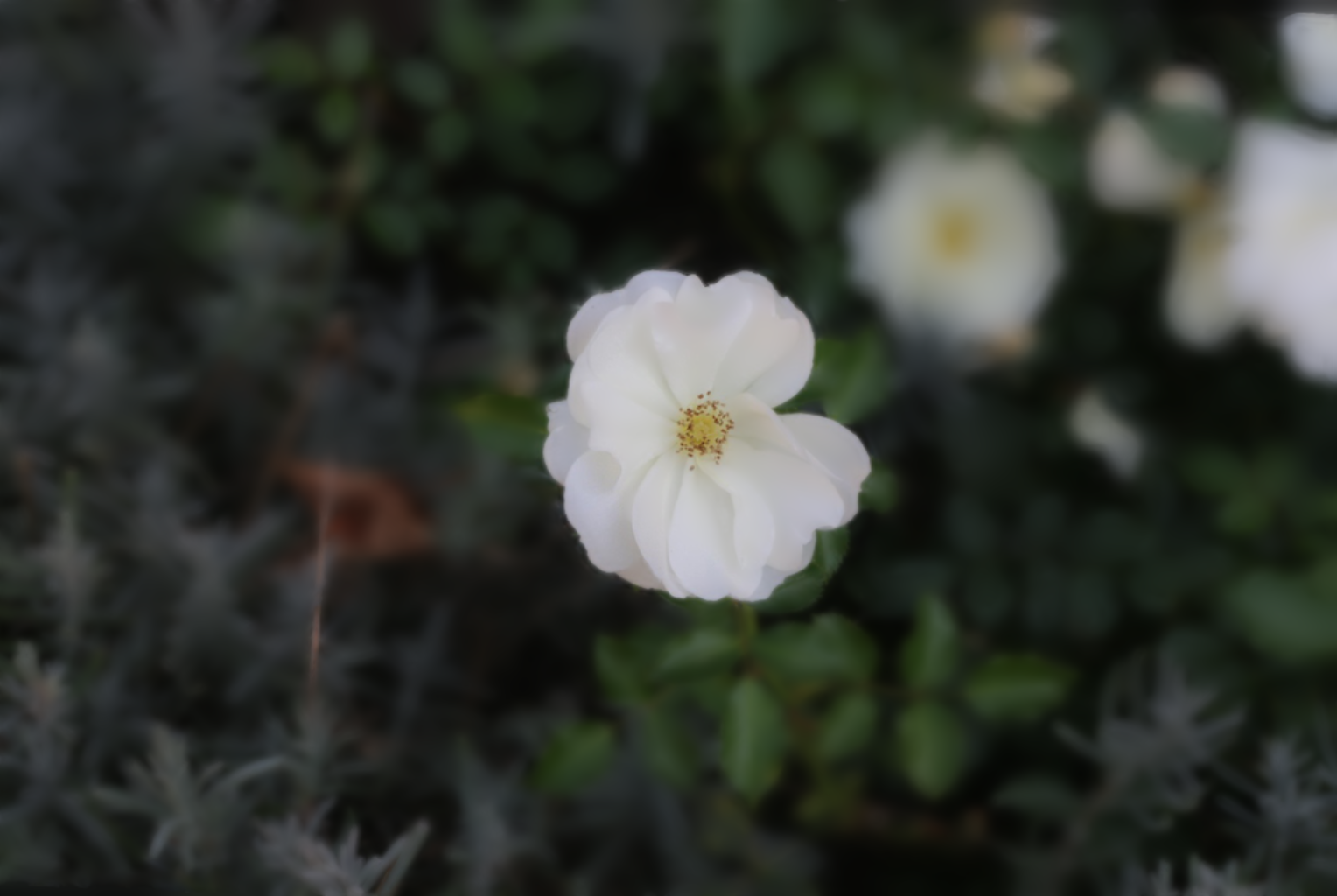}} \hfill
 \subfloat[Ground Truth]{\includegraphics[width = 0.24\linewidth,height=1.2in]{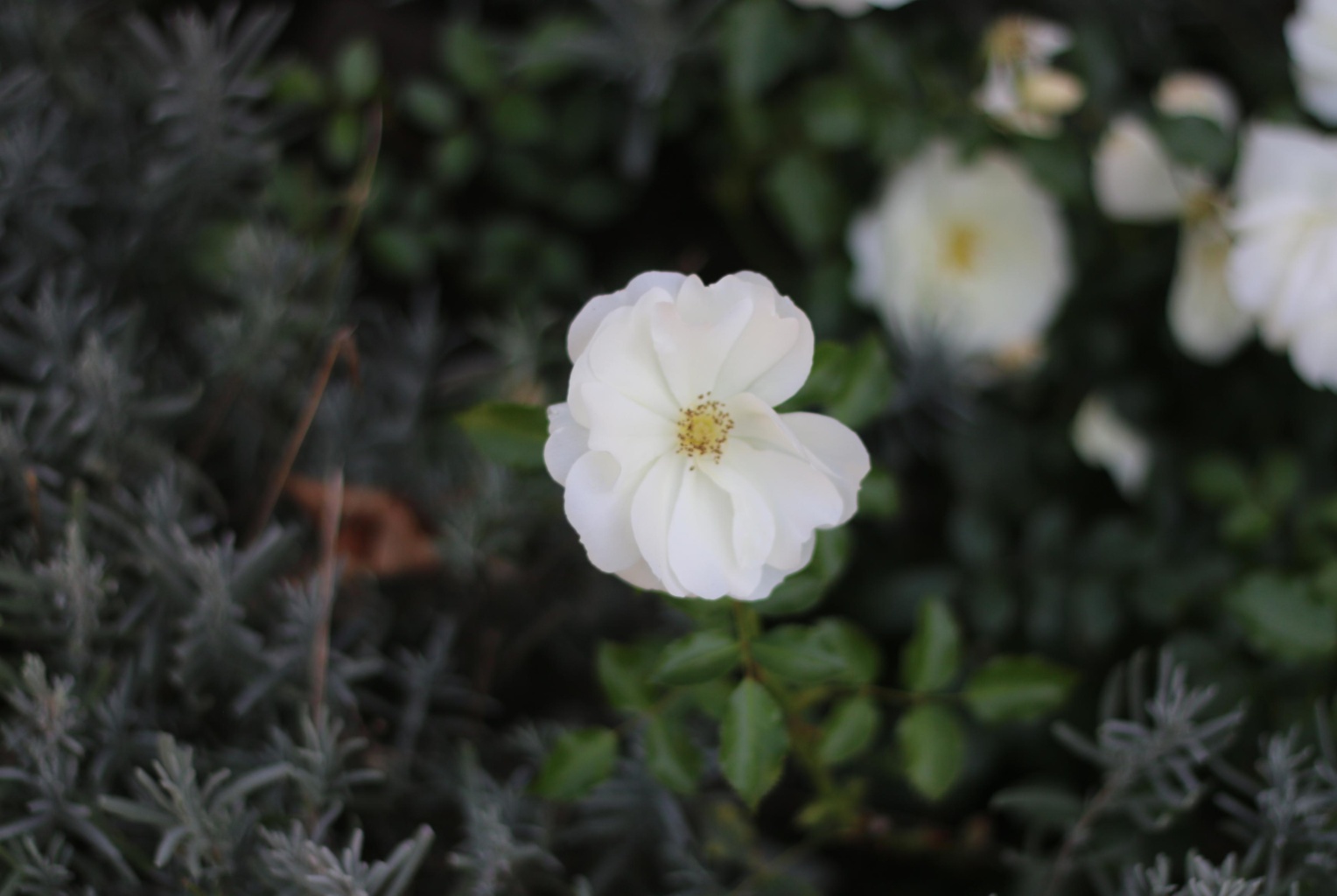}} \hfill \\
 
\subfloat[$W_3(k_3=45)$]{\includegraphics[width = 0.24\linewidth,height=1.2in]{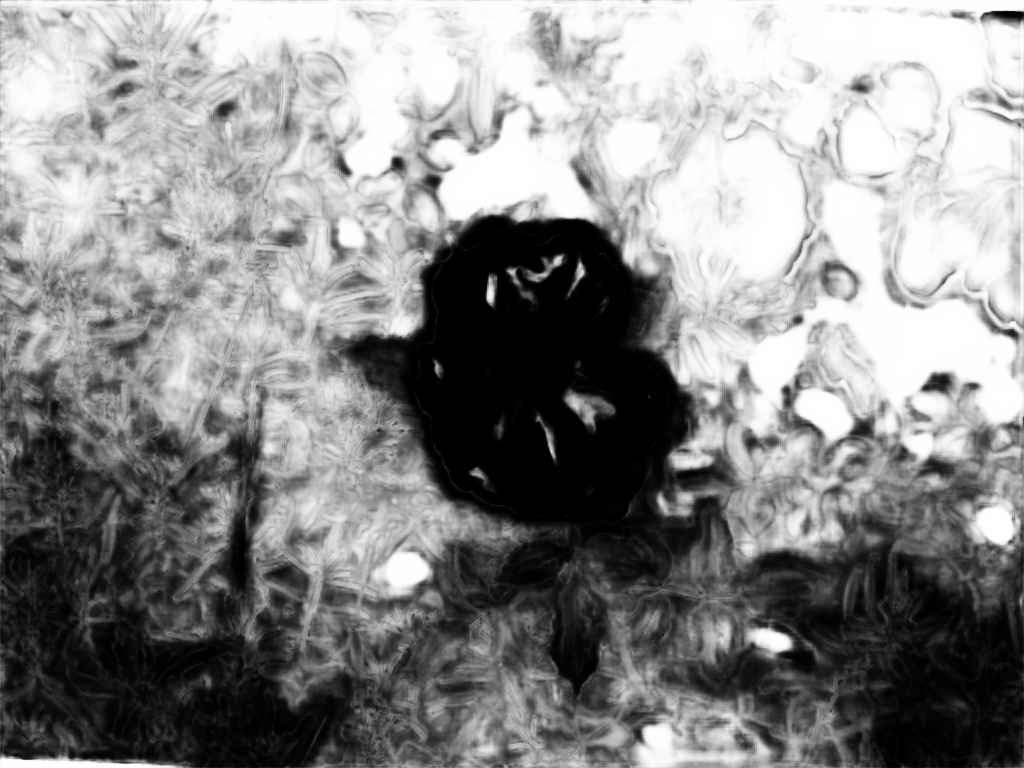}} \hfill
 \subfloat[$W_2(k_2=25)$]{\includegraphics[width = 0.24\linewidth,height=1.2in]{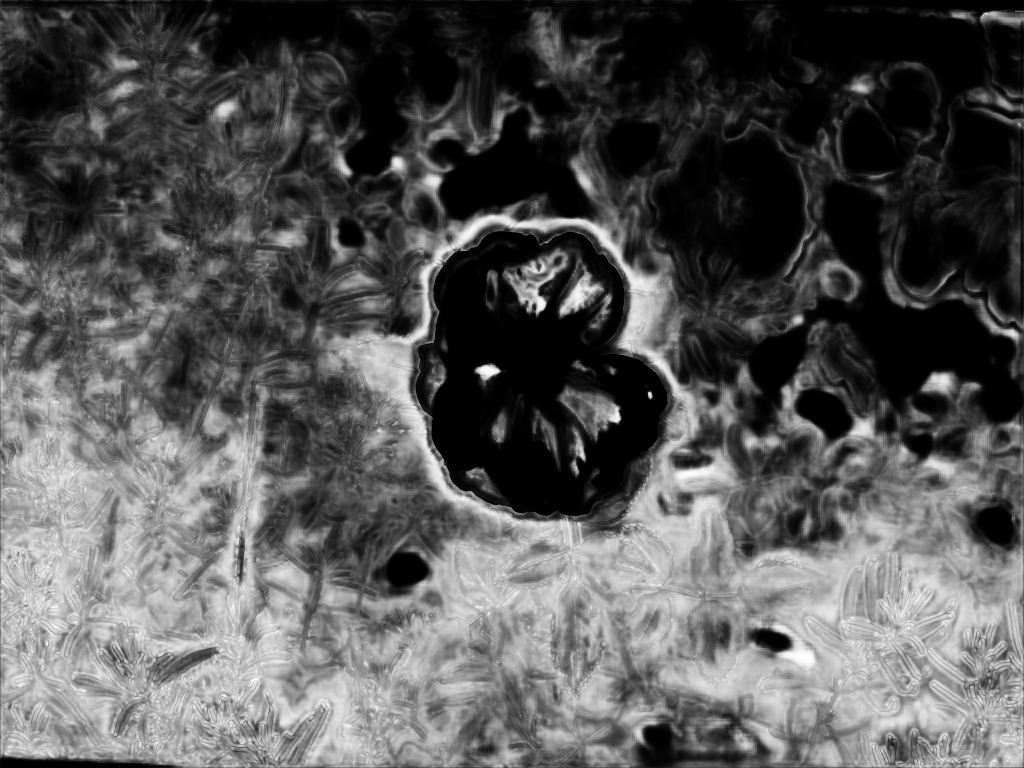}} \hfill
 \subfloat[$W_1(k_1=5)$]{\includegraphics[width = 0.24\linewidth,height=1.2in]{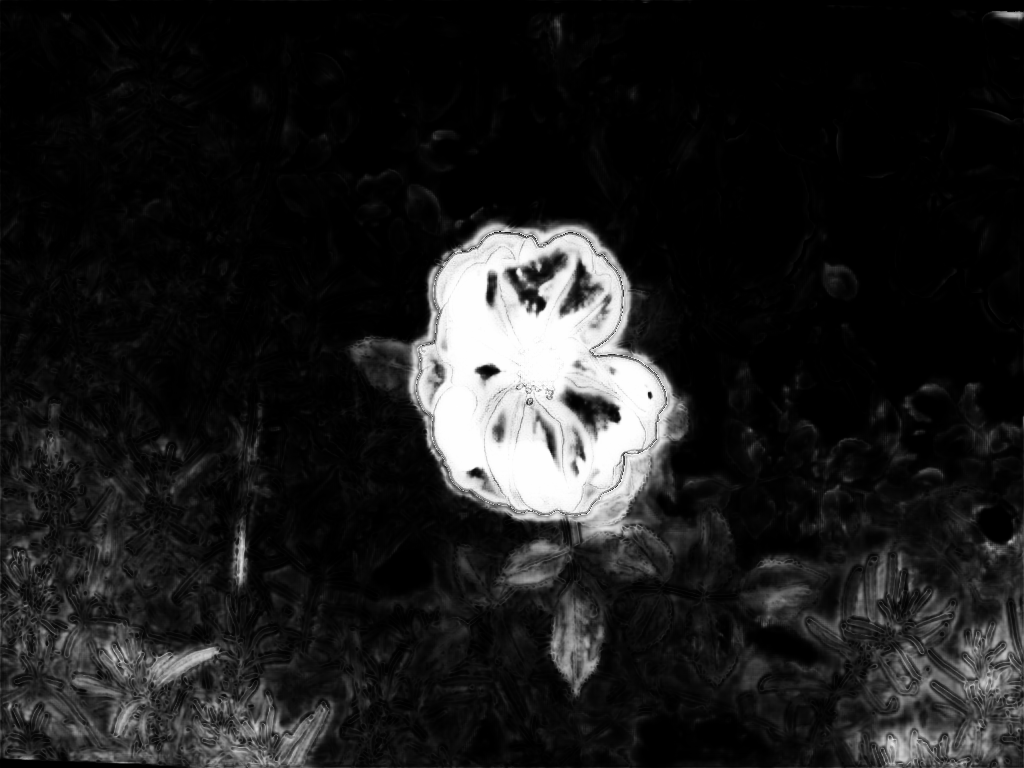}} \hfill
 \subfloat[$W_0$]{\includegraphics[width = 0.24\linewidth,height=1.2in]{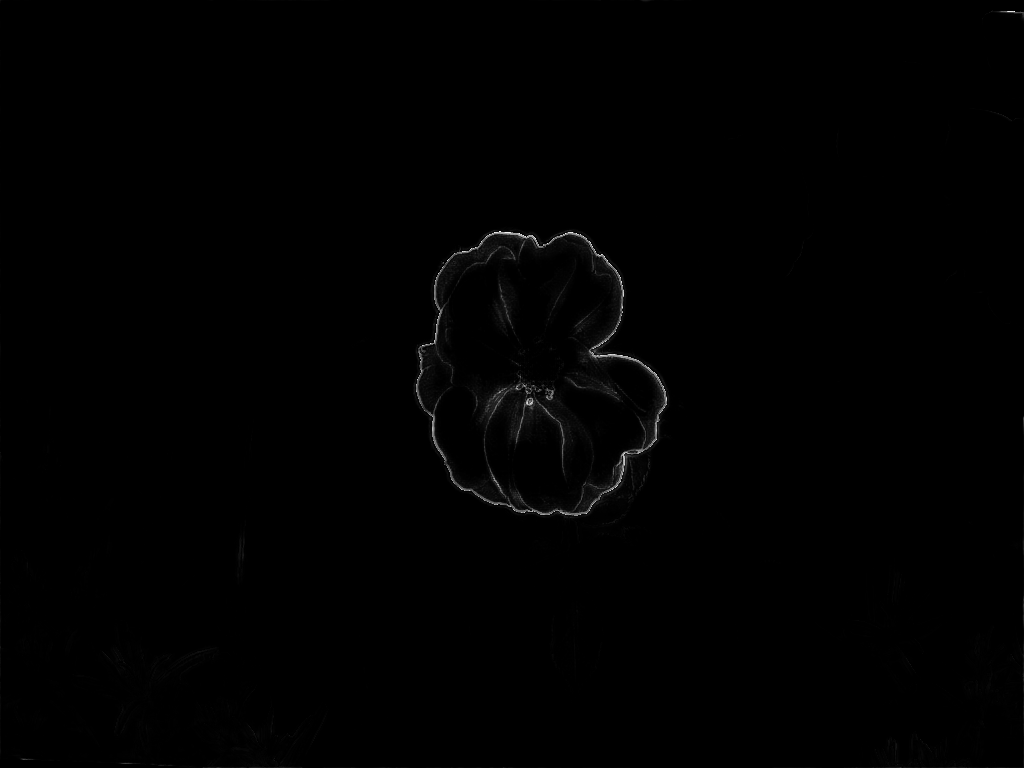}} \hfill \\ 
 
 \subfloat[$W_3(k_3=75)$]{\includegraphics[width = 0.24\linewidth,height=1.2in]{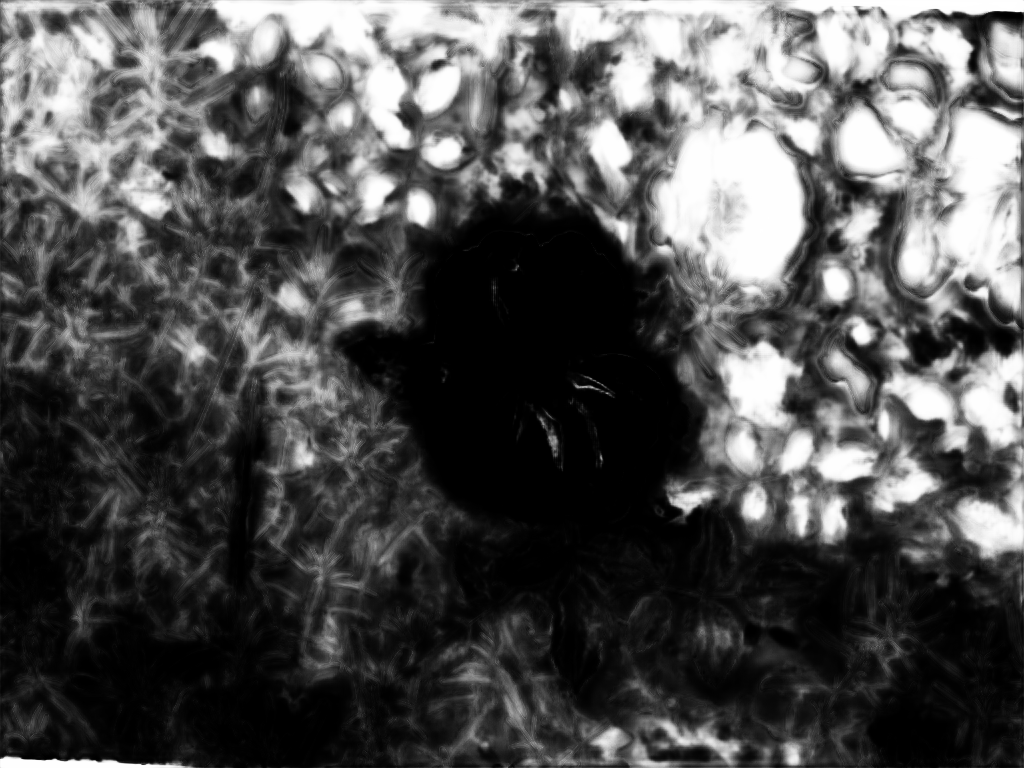}} \hfill
 \subfloat[$W_2(k_2=45)$]{\includegraphics[width = 0.24\linewidth,height=1.2in]{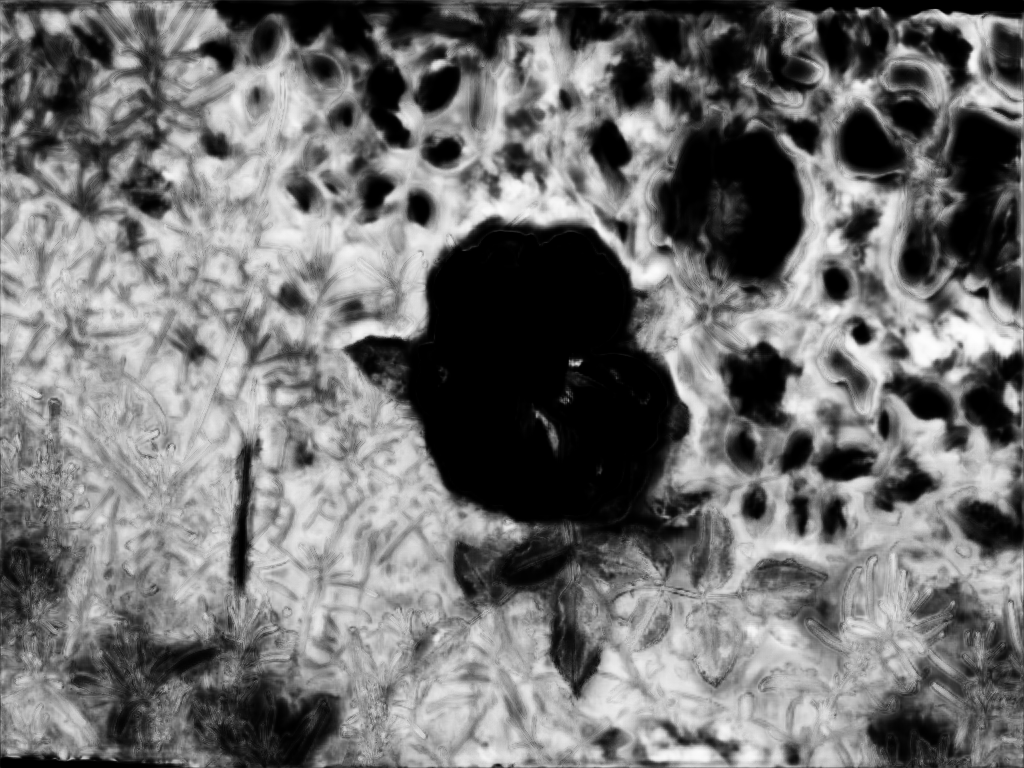}} \hfill
 \subfloat[$W_1(k_1=25)$]{\includegraphics[width = 0.24\linewidth,height=1.2in]{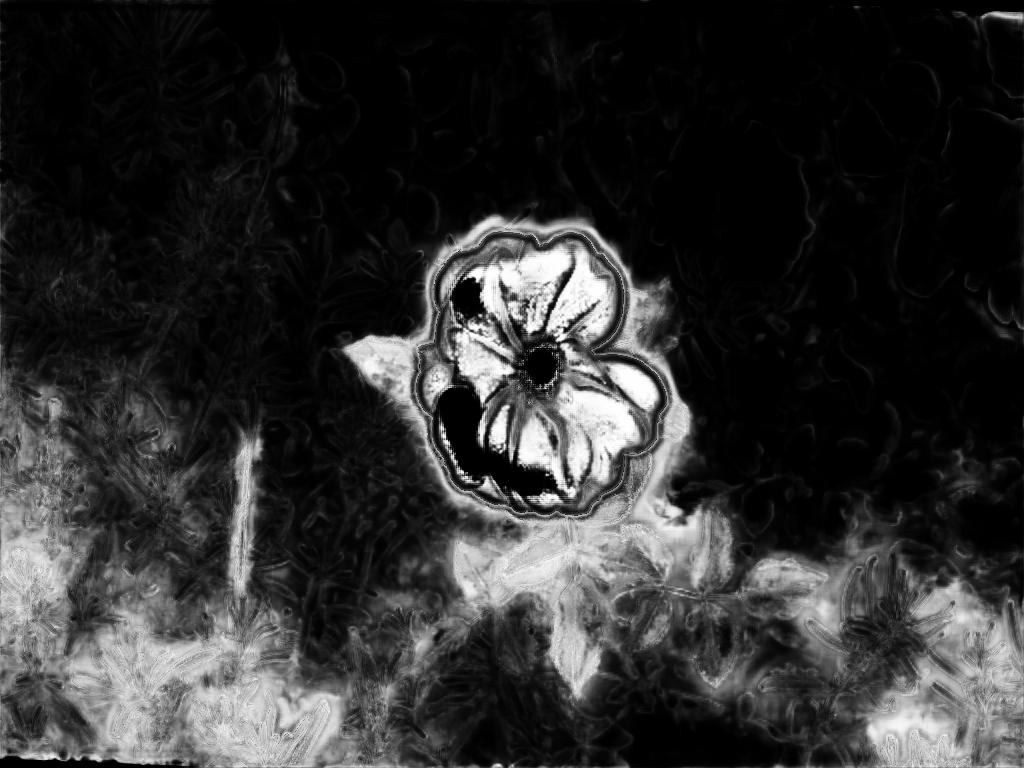}} \hfill
 \subfloat[$W_0$]{\includegraphics[width = 0.24\linewidth,height=1.2in]{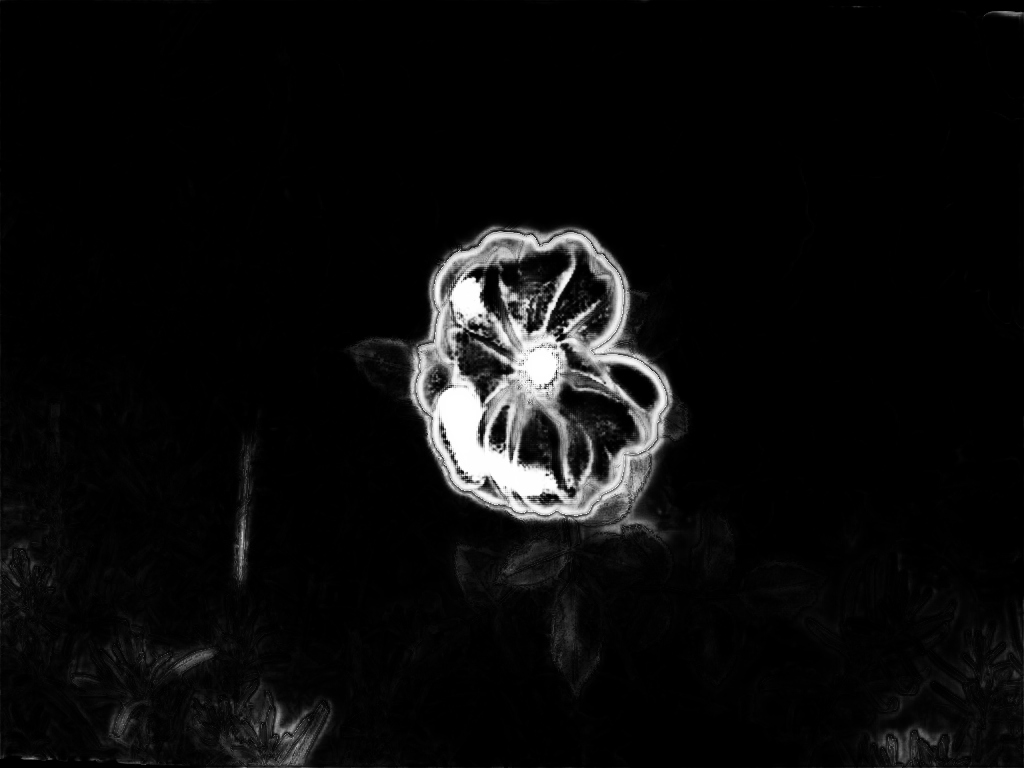}} \hfill \\ 
 
 \subfloat[Cropped from Bokeh using small kernels]{\includegraphics[width = 0.26\linewidth,height=1.4in]{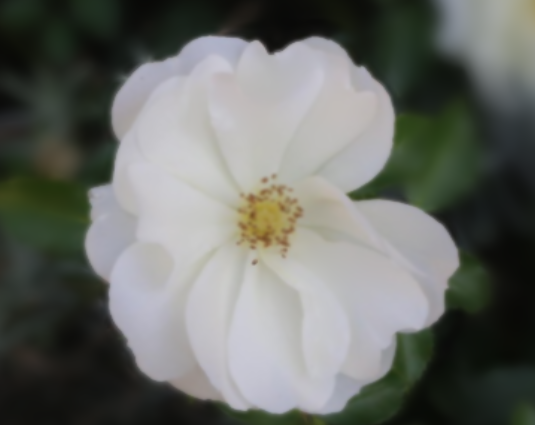}} \hfill
 \subfloat[Cropped from Bokeh using bigger kernels]{\includegraphics[width = 0.26\linewidth,height=1.4in]{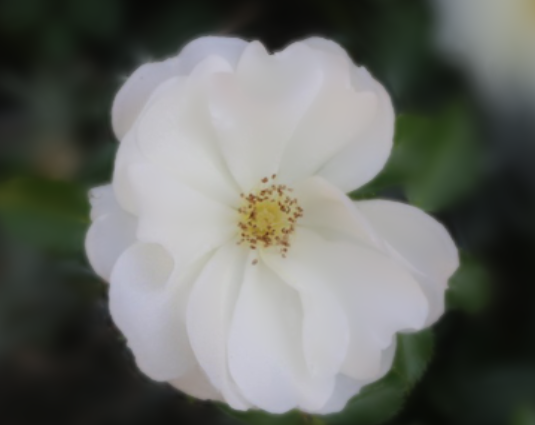}} \hfill
 \subfloat[Cropped from Ground Truth Bokeh Image]{\includegraphics[width = 0.26\linewidth,height=1.4in]{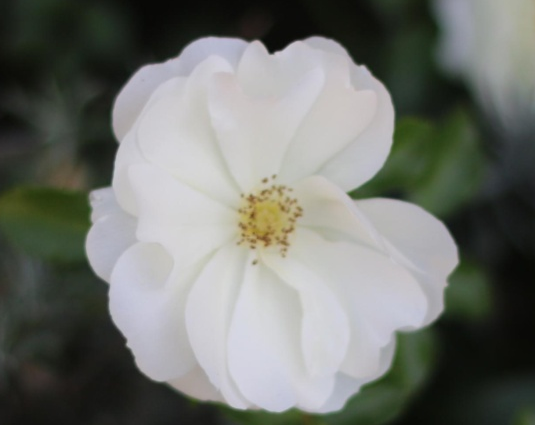}} \hfill
 
\caption{Qualitative comparison between Bokeh Images generated using smaller and bigger kernels. (e), (f), (g), (h) show weight maps using smaller kernels and (i), (j), (k), (l) show weight maps using bigger kernels.} 
\label{small_kernels}
\end{figure*}

\textbf{Smaller kernels vs bigger kernels.} To demonstrate the effect of using bigger kernels, two different sets of blur kernels were used. In one experiment, blur kernel sizes are respectively 5, 25 and 45 and in another experiment blur kernels of size 25, 45 and 75 are used. Table-\ref{tab_small_kernel} shows that using bigger kernels yields better performance. Qualitative comparison between these two settings are shown in Figure-\ref{small_kernels}. By analysing weight maps $W_1$ and $W_0$ in both settings, we observe that the value of $W_0$ is more in the in-focus region in case of bigger kernels than of smaller kernels. This helps network that uses bigger blur kernels to generate bokeh images that preserve more details in closer objects.


\begin{table}[!h]
\begin{tabular}{|c|c|c|c|}
\hline
                                                                                    & PSNR $\uparrow$ & SSIM $\uparrow$ & LPIPS $\downarrow$ \\ \hline
$k_1 = 5, k_2= 25, k_3= 45$                                                         &     23.22          &  0.8572             &    0.2607              \\ \hline
\begin{tabular}[c]{@{}c@{}}$k_1 = 25, k_2 = 45, k_3 = 75$\\ (Proposed)\end{tabular} &   \textbf{ 23.45}           &      \textbf{ 0.8675}      &    \textbf{0.2463}              \\ \hline
\end{tabular}
\centering
\caption{Quantitative comparison between bokeh effect generated by smaller kernels and bigger kernels on Val294 set.}
\label{tab_small_kernel}
\end{table}

\begin{figure*}[!h!t]
    \centering
    \includegraphics[width=\textwidth]{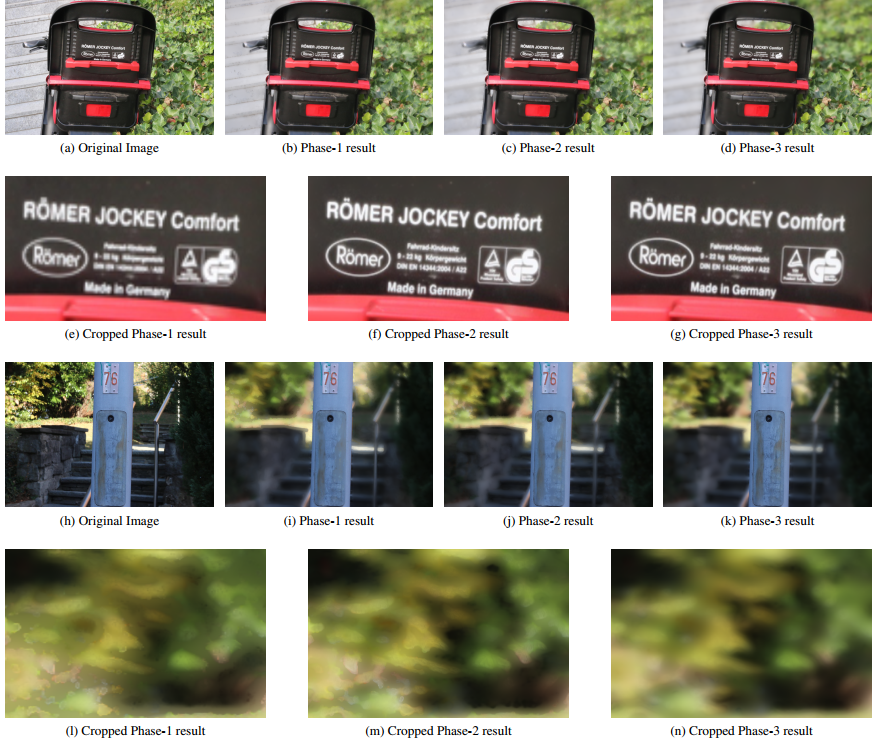}
    \caption{Qualitative Comparison between results at the end of three phases of training.}
    \label{3phase}
\end{figure*}

\textbf{Effect of three-phase training.}
The model is trained in three phases. As we can see in Table-\ref{tab_3phase}, results from Phase-1 scores the highest PSNR among all the phases, whereas Phase-3 produces best results with respect to SSIM and LPIPS. We observe in Figure-\ref{3phase} that Phase-2 and Phase-3 results preserve more details in the closer object than Phase-1 as Phase-2 and 3 models are trained on higher resolution images. Also, through fine-tuning with SSIM loss in Phase-3, the model generates a smoother blur effect in the background, as shown in Figure-\ref{3phase}.

\begin{table}[!h]
\begin{tabular}{|c|c|c|c|}
\hline
        & PSNR $\uparrow$ & SSIM $\uparrow$ & LPIPS $\downarrow$ \\ \hline
Phase-1 & \textbf{23.91}         & 0.8533        & 0.2716           \\ \hline
Phase-2 & 23.56         & 0.8557        & 0.2574           \\ \hline
Phase-3 & 23.45         & \textbf{0.8675}        & \textbf{0.2463}           \\ \hline
\end{tabular}
\centering
\caption{Quantitative Comparison among results from different phases on Val294 set.}
\label{tab_3phase}
\end{table}

\subsection{AIM 2019 Challenge on Bokeh Effect Synthesis}
The proposed solution participated in AIM 2019 Challenge on Bokeh Effect Synthesis which is a competition on example-based Bokeh effect generation. This competition had two tracks- namely Fidelity track and Perceptual track. In Fidelity track, submissions where judged based on PSNR and SSIM scores, whereas Mean Opinion Score(MOS) based on a user study was used to rank submissions in Perceptual track. 80 participants registered in this competition among which 9 teams entered the final phase. Two submissions were made to the separate tracks of the competition. Phase-1 results are submitted to Fidelity Track and Phase-3 results are submitted to Perceptual Track. The test results are shown in Table-\ref{aim_leaderboard}. The submission to Fidelity track (CVL-IITM-FDL) ranked 6th and 3rd among all the submissions with respect to PSNR and SSIM respectively. The Perceptual Track submission (CVL-IITM-PERC) ranked 2nd among all the submissions based on MOS score.

\begin{table}[t!]
\begin{tabular}{|c|c|c|c|}
\hline
    Team            & PSNR $\uparrow$  & SSIM $\uparrow$ & MOS $\downarrow$  \\ \hline
islab-zju       & 23.43 & 0.89 & 0.87 \\ \hline
CVL-IITM-PERC   & 22.14 & 0.86 & 0.88 \\ \hline
\begin{tabular}[c]{@{}c@{}}IPCV IITM\\ \citep{purohit2019depth}\end{tabular}
      & 23.17 & 0.88 & 0.93 \\ \hline
VIDAR           & 23.93 & 0.89 & 0.98 \\ \hline
XMU-VIPLab      & 23.18 & 0.89 & 1.02 \\ \hline
KAIST-VICLAB    & 23.37 & 0.86 & 1.08 \\ \hline
SLYYM           & 22.25 & 0.85 & 1.10 \\ \hline
Bazinga         & 22.15 & 0.84 & 1.48 \\ \hline
aps vision@IITG & 20.88 & 0.77 & 1.53 \\ \hline
CVL-IITM-FDL    & 22.90 & 0.87 & N/A  \\ \hline
\end{tabular}
\centering
\caption{AIM 2019 Challenge on Bokeh Effect Synthesis Leaderboard.}
\label{aim_leaderboard}
\end{table}

\subsection{Efficiency} The proposed model is lightweight. It takes 28 MB to store the model on hard drive. The model can process an HD image of resolution $1024 \times 1575$ in $0.03$ seconds (I/O time is excluded).

\section{Conclusion}
In this paper, an end-to-end deep learning approach for Bokeh effect synthesis is proposed. The synthesized bokeh image is rendered as a weighted sum of the input image and a number of differently smoothed images, where the corresponding weight maps are predicted by a depth-estimation network. The proposed system is trained in three phases to synthesize realistic bokeh images. It is shown through experiments that using more number of blur kernels and bigger blur kernels produce better quality bokeh images. The proposed algorithm is lightweight and can post-process an HD image in 0.03 seconds. The proposed approach has been ranked 2nd among the solutions proposed in Perceptual Track of AIM 2019 challenge on Bokeh effect synthesis. In future, the effect of using different kinds of blur kernel (e.g. average blur, disk blur) can be explored. It is also interesting to see how one can incorporate both monocular depth estimation and saliency detection to produce a lightweight system that generates high-quality bokeh effect.

\bibliographystyle{model2-names}
\bibliography{refs}

\begin{thebibliography}{22}
\expandafter\ifx\csname natexlab\endcsname\relax\def\natexlab#1{#1}\fi
\providecommand{\url}[1]{\texttt{#1}}
\providecommand{\href}[2]{#2}
\providecommand{\path}[1]{#1}
\providecommand{\DOIprefix}{doi:}
\providecommand{\ArXivprefix}{arXiv:}
\providecommand{\URLprefix}{URL: }
\providecommand{\Pubmedprefix}{pmid:}
\providecommand{\doi}[1]{\href{http://dx.doi.org/#1}{\path{#1}}}
\providecommand{\Pubmed}[1]{\href{pmid:#1}{\path{#1}}}
\providecommand{\bibinfo}[2]{#2}
\ifx\xfnm\relax \def\xfnm[#1]{\unskip,\space#1}\fi
\bibitem[{Atapour-Abarghouei and Breckon(2018)}]{depth_style_transfer}
\bibinfo{author}{Atapour-Abarghouei, A.}, \bibinfo{author}{Breckon, T.P.},
  \bibinfo{year}{2018}.
\newblock \bibinfo{title}{Real-time monocular depth estimation using synthetic
  data with domain adaptation via image style transfer}, in:
  \bibinfo{booktitle}{Proceedings of the IEEE Conference on Computer Vision and
  Pattern Recognition}, pp. \bibinfo{pages}{2800--2810}.
\bibitem[{Chen et~al.(2016)Chen, Fu, Yang and Deng}]{chen2016single}
\bibinfo{author}{Chen, W.}, \bibinfo{author}{Fu, Z.}, \bibinfo{author}{Yang,
  D.}, \bibinfo{author}{Deng, J.}, \bibinfo{year}{2016}.
\newblock \bibinfo{title}{Single-image depth perception in the wild}, in:
  \bibinfo{booktitle}{Advances in neural information processing systems}, pp.
  \bibinfo{pages}{730--738}.
\bibitem[{Geiger et~al.(2012)Geiger, Lenz and Urtasun}]{geiger2012we}
\bibinfo{author}{Geiger, A.}, \bibinfo{author}{Lenz, P.},
  \bibinfo{author}{Urtasun, R.}, \bibinfo{year}{2012}.
\newblock \bibinfo{title}{Are we ready for autonomous driving? the kitti vision
  benchmark suite}, in: \bibinfo{booktitle}{2012 IEEE Conference on Computer
  Vision and Pattern Recognition}, \bibinfo{organization}{IEEE}. pp.
  \bibinfo{pages}{3354--3361}.
\bibitem[{Godard et~al.(2017)Godard, Mac~Aodha and Brostow}]{monodepth}
\bibinfo{author}{Godard, C.}, \bibinfo{author}{Mac~Aodha, O.},
  \bibinfo{author}{Brostow, G.J.}, \bibinfo{year}{2017}.
\newblock \bibinfo{title}{Unsupervised monocular depth estimation with
  left-right consistency}, in: \bibinfo{booktitle}{Proceedings of the IEEE
  Conference on Computer Vision and Pattern Recognition}, pp.
  \bibinfo{pages}{270--279}.
\bibitem[{Godard et~al.(2019)Godard, Mac~Aodha, Firman and
  Brostow}]{monodepth2}
\bibinfo{author}{Godard, C.}, \bibinfo{author}{Mac~Aodha, O.},
  \bibinfo{author}{Firman, M.}, \bibinfo{author}{Brostow, G.J.},
  \bibinfo{year}{2019}.
\newblock \bibinfo{title}{Digging into self-supervised monocular depth
  estimation}, in: \bibinfo{booktitle}{Proceedings of the IEEE International
  Conference on Computer Vision}, pp. \bibinfo{pages}{3828--3838}.
\bibitem[{Ignatov et~al.(2019)Ignatov, Patel, Timofte, Zheng, Ye, Huang, Tian,
  Dutta, Purohit, Kandula et~al.}]{ignatov2019aim}
\bibinfo{author}{Ignatov, A.}, \bibinfo{author}{Patel, J.},
  \bibinfo{author}{Timofte, R.}, \bibinfo{author}{Zheng, B.},
  \bibinfo{author}{Ye, X.}, \bibinfo{author}{Huang, L.}, \bibinfo{author}{Tian,
  X.}, \bibinfo{author}{Dutta, S.}, \bibinfo{author}{Purohit, K.},
  \bibinfo{author}{Kandula, P.}, et~al., \bibinfo{year}{2019}.
\newblock \bibinfo{title}{Aim 2019 challenge on bokeh effect synthesis: Methods
  and results}, in: \bibinfo{booktitle}{2019 IEEE/CVF International Conference
  on Computer Vision Workshop (ICCVW)}, \bibinfo{organization}{IEEE}. pp.
  \bibinfo{pages}{3591--3598}.
\bibitem[{Kingma and Ba(2014)}]{kingma2014adam}
\bibinfo{author}{Kingma, D.P.}, \bibinfo{author}{Ba, J.}, \bibinfo{year}{2014}.
\newblock \bibinfo{title}{Adam: A method for stochastic optimization}.
\newblock \bibinfo{journal}{arXiv preprint arXiv:1412.6980} .
\bibitem[{Li and Snavely(2018)}]{li2018megadepth}
\bibinfo{author}{Li, Z.}, \bibinfo{author}{Snavely, N.}, \bibinfo{year}{2018}.
\newblock \bibinfo{title}{Megadepth: Learning single-view depth prediction from
  internet photos}, in: \bibinfo{booktitle}{Proceedings of the IEEE Conference
  on Computer Vision and Pattern Recognition}, pp. \bibinfo{pages}{2041--2050}.
\bibitem[{Liu et~al.(2017)Liu, De~Mello, Gu, Zhong, Yang and
  Kautz}]{liu2017learning}
\bibinfo{author}{Liu, S.}, \bibinfo{author}{De~Mello, S.}, \bibinfo{author}{Gu,
  J.}, \bibinfo{author}{Zhong, G.}, \bibinfo{author}{Yang, M.H.},
  \bibinfo{author}{Kautz, J.}, \bibinfo{year}{2017}.
\newblock \bibinfo{title}{Learning affinity via spatial propagation networks},
  in: \bibinfo{booktitle}{Advances in Neural Information Processing Systems},
  pp. \bibinfo{pages}{1520--1530}.
\bibitem[{Liu et~al.(2016)Liu, Pan and Yang}]{liu2016learning}
\bibinfo{author}{Liu, S.}, \bibinfo{author}{Pan, J.}, \bibinfo{author}{Yang,
  M.H.}, \bibinfo{year}{2016}.
\newblock \bibinfo{title}{Learning recursive filters for low-level vision via a
  hybrid neural network}, in: \bibinfo{booktitle}{European Conference on
  Computer Vision}, \bibinfo{organization}{Springer}. pp.
  \bibinfo{pages}{560--576}.
\bibitem[{Long et~al.(2015)Long, Shelhamer and Darrell}]{fcn}
\bibinfo{author}{Long, J.}, \bibinfo{author}{Shelhamer, E.},
  \bibinfo{author}{Darrell, T.}, \bibinfo{year}{2015}.
\newblock \bibinfo{title}{Fully convolutional networks for semantic
  segmentation}, in: \bibinfo{booktitle}{Proceedings of the IEEE conference on
  computer vision and pattern recognition}, pp. \bibinfo{pages}{3431--3440}.
\bibitem[{Luo et~al.(2018)Luo, Ren, Lin, Pang, Sun, Li and Lin}]{svs}
\bibinfo{author}{Luo, Y.}, \bibinfo{author}{Ren, J.}, \bibinfo{author}{Lin,
  M.}, \bibinfo{author}{Pang, J.}, \bibinfo{author}{Sun, W.},
  \bibinfo{author}{Li, H.}, \bibinfo{author}{Lin, L.}, \bibinfo{year}{2018}.
\newblock \bibinfo{title}{Single view stereo matching}, in:
  \bibinfo{booktitle}{Proceedings of the IEEE Conference on Computer Vision and
  Pattern Recognition}, pp. \bibinfo{pages}{155--163}.
\bibitem[{Paszke et~al.(2019)Paszke, Gross, Massa, Lerer, Bradbury, Chanan,
  Killeen, Lin, Gimelshein, Antiga, Desmaison, Kopf, Yang, DeVito, Raison,
  Tejani, Chilamkurthy, Steiner, Fang, Bai and Chintala}]{pytorch}
\bibinfo{author}{Paszke, A.}, \bibinfo{author}{Gross, S.},
  \bibinfo{author}{Massa, F.}, \bibinfo{author}{Lerer, A.},
  \bibinfo{author}{Bradbury, J.}, \bibinfo{author}{Chanan, G.},
  \bibinfo{author}{Killeen, T.}, \bibinfo{author}{Lin, Z.},
  \bibinfo{author}{Gimelshein, N.}, \bibinfo{author}{Antiga, L.},
  \bibinfo{author}{Desmaison, A.}, \bibinfo{author}{Kopf, A.},
  \bibinfo{author}{Yang, E.}, \bibinfo{author}{DeVito, Z.},
  \bibinfo{author}{Raison, M.}, \bibinfo{author}{Tejani, A.},
  \bibinfo{author}{Chilamkurthy, S.}, \bibinfo{author}{Steiner, B.},
  \bibinfo{author}{Fang, L.}, \bibinfo{author}{Bai, J.},
  \bibinfo{author}{Chintala, S.}, \bibinfo{year}{2019}.
\newblock \bibinfo{title}{Pytorch: An imperative style, high-performance deep
  learning library}, in: \bibinfo{booktitle}{Advances in Neural Information
  Processing Systems 32}, pp. \bibinfo{pages}{8024--8035}.
\bibitem[{Purohit et~al.(2019)Purohit, Suin, Kandula and
  Ambasamudram}]{purohit2019depth}
\bibinfo{author}{Purohit, K.}, \bibinfo{author}{Suin, M.},
  \bibinfo{author}{Kandula, P.}, \bibinfo{author}{Ambasamudram, R.},
  \bibinfo{year}{2019}.
\newblock \bibinfo{title}{Depth-guided dense dynamic filtering network for
  bokeh effect rendering}, in: \bibinfo{booktitle}{2019 IEEE/CVF International
  Conference on Computer Vision Workshop (ICCVW)},
  \bibinfo{organization}{IEEE}. pp. \bibinfo{pages}{3417--3426}.
\bibitem[{Saxena et~al.(2006)Saxena, Chung and Ng}]{saxena2006learning}
\bibinfo{author}{Saxena, A.}, \bibinfo{author}{Chung, S.H.},
  \bibinfo{author}{Ng, A.Y.}, \bibinfo{year}{2006}.
\newblock \bibinfo{title}{Learning depth from single monocular images}, in:
  \bibinfo{booktitle}{Advances in neural information processing systems}, pp.
  \bibinfo{pages}{1161--1168}.
\bibitem[{Shen et~al.(2016)Shen, Hertzmann, Jia, Paris, Price, Shechtman and
  Sachs}]{shen2016automatic}
\bibinfo{author}{Shen, X.}, \bibinfo{author}{Hertzmann, A.},
  \bibinfo{author}{Jia, J.}, \bibinfo{author}{Paris, S.},
  \bibinfo{author}{Price, B.}, \bibinfo{author}{Shechtman, E.},
  \bibinfo{author}{Sachs, I.}, \bibinfo{year}{2016}.
\newblock \bibinfo{title}{Automatic portrait segmentation for image
  stylization}, in: \bibinfo{booktitle}{Computer Graphics Forum},
  \bibinfo{organization}{Wiley Online Library}. pp. \bibinfo{pages}{93--102}.
\bibitem[{Wadhwa et~al.(2018)Wadhwa, Garg, Jacobs, Feldman, Kanazawa, Carroll,
  Movshovitz-Attias, Barron, Pritch and Levoy}]{wadhwa2018synthetic}
\bibinfo{author}{Wadhwa, N.}, \bibinfo{author}{Garg, R.},
  \bibinfo{author}{Jacobs, D.E.}, \bibinfo{author}{Feldman, B.E.},
  \bibinfo{author}{Kanazawa, N.}, \bibinfo{author}{Carroll, R.},
  \bibinfo{author}{Movshovitz-Attias, Y.}, \bibinfo{author}{Barron, J.T.},
  \bibinfo{author}{Pritch, Y.}, \bibinfo{author}{Levoy, M.},
  \bibinfo{year}{2018}.
\newblock \bibinfo{title}{Synthetic depth-of-field with a single-camera mobile
  phone}.
\newblock \bibinfo{journal}{ACM Transactions on Graphics (TOG)}
  \bibinfo{volume}{37}, \bibinfo{pages}{1--13}.
\bibitem[{Wang et~al.(2004)Wang, Bovik, Sheikh and Simoncelli}]{ssim_paper}
\bibinfo{author}{Wang, Z.}, \bibinfo{author}{Bovik, A.C.},
  \bibinfo{author}{Sheikh, H.R.}, \bibinfo{author}{Simoncelli, E.P.},
  \bibinfo{year}{2004}.
\newblock \bibinfo{title}{Image quality assessment: from error visibility to
  structural similarity}.
\newblock \bibinfo{journal}{IEEE transactions on image processing}
  \bibinfo{volume}{13}, \bibinfo{pages}{600--612}.
\bibitem[{Wu et~al.(2019)Wu, Su and Huang}]{wu2019stacked}
\bibinfo{author}{Wu, Z.}, \bibinfo{author}{Su, L.}, \bibinfo{author}{Huang,
  Q.}, \bibinfo{year}{2019}.
\newblock \bibinfo{title}{Stacked cross refinement network for edge-aware
  salient object detection}, in: \bibinfo{booktitle}{Proceedings of the IEEE
  International Conference on Computer Vision}, pp.
  \bibinfo{pages}{7264--7273}.
\bibitem[{Xu et~al.(2018)Xu, Sun, Liu, Ren, Zhang, Yang and
  Sun}]{xu2018rendering}
\bibinfo{author}{Xu, X.}, \bibinfo{author}{Sun, D.}, \bibinfo{author}{Liu, S.},
  \bibinfo{author}{Ren, W.}, \bibinfo{author}{Zhang, Y.J.},
  \bibinfo{author}{Yang, M.H.}, \bibinfo{author}{Sun, J.},
  \bibinfo{year}{2018}.
\newblock \bibinfo{title}{Rendering portraitures from monocular camera and
  beyond}, in: \bibinfo{booktitle}{Proceedings of the European Conference on
  Computer Vision (ECCV)}, pp. \bibinfo{pages}{35--50}.
\bibitem[{Zhang et~al.(2018)Zhang, Isola, Efros, Shechtman and Wang}]{LPIPS}
\bibinfo{author}{Zhang, R.}, \bibinfo{author}{Isola, P.},
  \bibinfo{author}{Efros, A.A.}, \bibinfo{author}{Shechtman, E.},
  \bibinfo{author}{Wang, O.}, \bibinfo{year}{2018}.
\newblock \bibinfo{title}{The unreasonable effectiveness of deep features as a
  perceptual metric}, in: \bibinfo{booktitle}{CVPR}, pp.
  \bibinfo{pages}{586--595}.
\bibitem[{Zheng et~al.(2015)Zheng, Jayasumana, Romera-Paredes, Vineet, Su, Du,
  Huang and Torr}]{zheng2015conditional}
\bibinfo{author}{Zheng, S.}, \bibinfo{author}{Jayasumana, S.},
  \bibinfo{author}{Romera-Paredes, B.}, \bibinfo{author}{Vineet, V.},
  \bibinfo{author}{Su, Z.}, \bibinfo{author}{Du, D.}, \bibinfo{author}{Huang,
  C.}, \bibinfo{author}{Torr, P.H.}, \bibinfo{year}{2015}.
\newblock \bibinfo{title}{Conditional random fields as recurrent neural
  networks}, in: \bibinfo{booktitle}{Proceedings of the IEEE international
  conference on computer vision}, pp. \bibinfo{pages}{1529--1537}.

\end{thebibliography}

\end{document}